\documentclass[runningheads]{llncs}

\PassOptionsToPackage{table}{xcolor}
\usepackage{eccv}

\usepackage{eccvabbrv}

\usepackage{graphicx}
\usepackage{booktabs}

\usepackage[accsupp]{axessibility}  %

\usepackage{hyperref}

\usepackage{orcidlink}

\usepackage{multirow}
\usepackage{anyfontsize}

\usepackage{overpic}
\usepackage{subcaption}
\usepackage{tabularx}
\usepackage{booktabs}
\usepackage{tabularx}
\usepackage{multirow}
\usepackage{pifont}   %
\usepackage{array}
\usepackage{comment}
\usepackage{wrapfig}
\usepackage{amssymb}

\newcommand{\cmark}{\textcolor{green!60!black}{$\checkmark$}}
\newcommand{\xmark}{\textcolor{red!70!black}{$\times$}}

\begin{document}

\title{OV3D-Bench: A Diagnostic Benchmark for Open-Vocabulary Monocular 3D Detection}

\titlerunning{A Diagnostic Benchmark for Open-Vocabulary Monocular 3D Detection}

\author{Mariia Gladkova$^1$, Neehar Peri$^2$, Ishan Khatri$^3$, Deva Ramanan$^2$, Daniel Cremers$^1$}

\authorrunning{Gladkova et al.}

\institute{Technical University of Munich \and Carnegie Mellon University \and StackAV}

\maketitle

\begin{abstract}
Open-vocabulary monocular 3D detectors report strong in-domain performance, but each evaluates under a different protocol, several rely on per-image category oracles unavailable at deployment, and all collapse geometry and semantics into a single AP metric. To address this, we introduce OV3D-Bench, a diagnostic benchmark that compares open-vocabulary monocular 3D detectors under deployment-realistic conditions across seven indoor and outdoor datasets. Our benchmark replaces the per-image class name oracle with test-time dataset-level class name prompts, and decouples detection accuracy along three axes: localization, semantic robustness, and cross-domain transfer. We evaluate seven representative detectors and find that (i) they localize objects well yet often mislabel a correctly localized box as a semantically adjacent category; (ii) accuracy is highly sensitive to prompt phrasing (\eg WildDet3D's performance collapses from 18.6 to 5.4 AP\textsubscript{3D} when prompted with ``a detailed high-resolution photo of a
car'' rather than  ``car''); and (iii) the widely adopted target-aware protocol hides these errors (\eg inflating DetAny3D's AP\textsubscript{3D} by 1.9 $\times$ on ScanNet). Lastly, we demonstrate that simply remapping a frozen {\em closed-vocabulary} detector's predictions using a contrastive vision-language encoder such as SigLIPv2 performs competitively against recent purpose-built open-vocabulary methods. This indicates that geometric localization is more mature, while open-vocabulary semantics remains the primary bottleneck. We release our unified evaluation protocol on \href{https://github.com/mgladkova/ov3d-bench}{GitHub}.
\end{abstract}

\section{Introduction}
\begin{figure*}[t]
\centering
\vspace{0.5em}
\begin{overpic}[width=\linewidth]{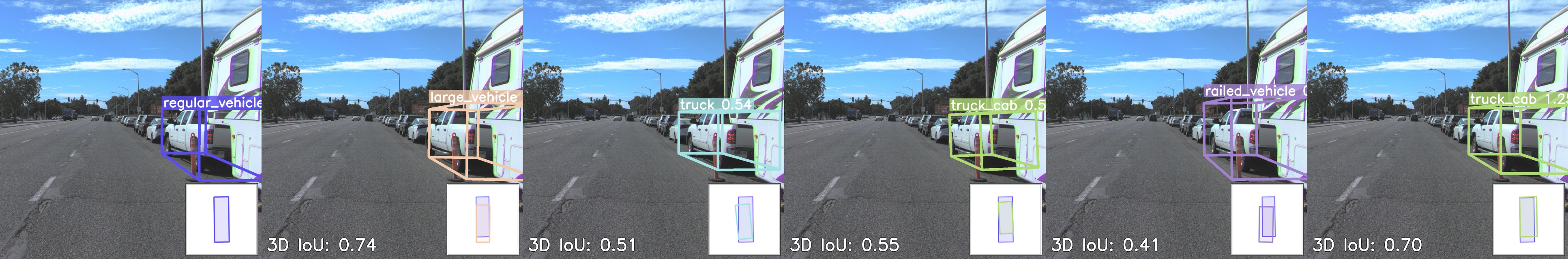}
    \put(1.0,17.5){\scriptsize Ground Truth}
    \put(18,17.5){\scriptsize Cube R-CNN*}
    \put(34.5,17.5){\scriptsize OVM3D-Det*}
    \put(53,17.5){\scriptsize DetAny3D}
    \put(69,17.5){\scriptsize OVMono3D}
    \put(86,17.5){\scriptsize WildDet3D}
    \put(-2.0,6.0){\rotatebox{90}{\scriptsize AV2}}
\end{overpic}\\[0.5em]
\begin{minipage}[b]{0.49\linewidth}
    \centering
    \includegraphics[width=\linewidth]{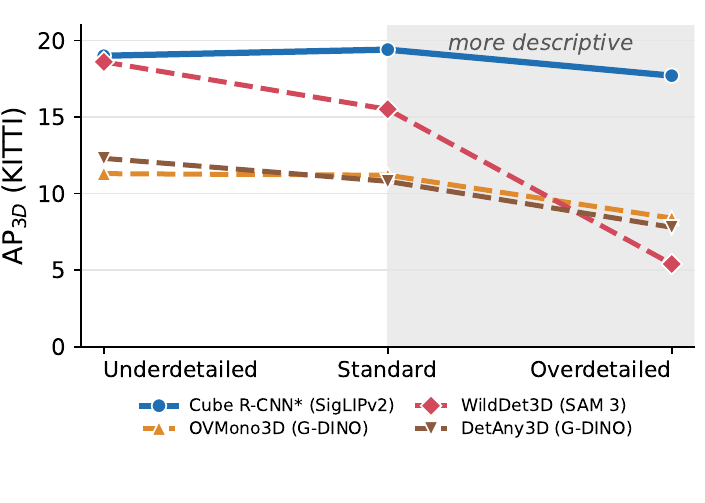}\\[-0.1em]
\end{minipage}\hfill
\begin{minipage}[b]{0.49\linewidth}
    \centering
    \includegraphics[width=\linewidth]{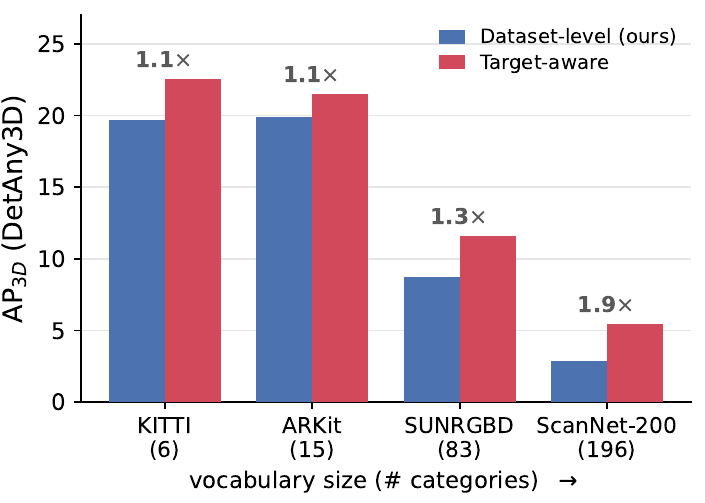}\\[-0.1em]
\end{minipage}
\caption{\textbf{Open-Vocabulary Monocular 3D Detectors Struggle with Vocabulary, Not Geometry.} All five detectors consistently localize a \texttt{regular vehicle} (mean 3D IoU $0.58$, BEV insets) yet disagree on its category \emph{(top)}. Prompt phrasing swings accuracy sharply for prompt-conditioned detectors but far less for our SigLIPv2-based remapping baseline \emph{(bottom left)}. The target-aware protocol hides these semantic errors, inflating scores most where the dataset-level score is lowest. \emph{(bottom right)}.}
\label{fig:teaser}
\end{figure*}
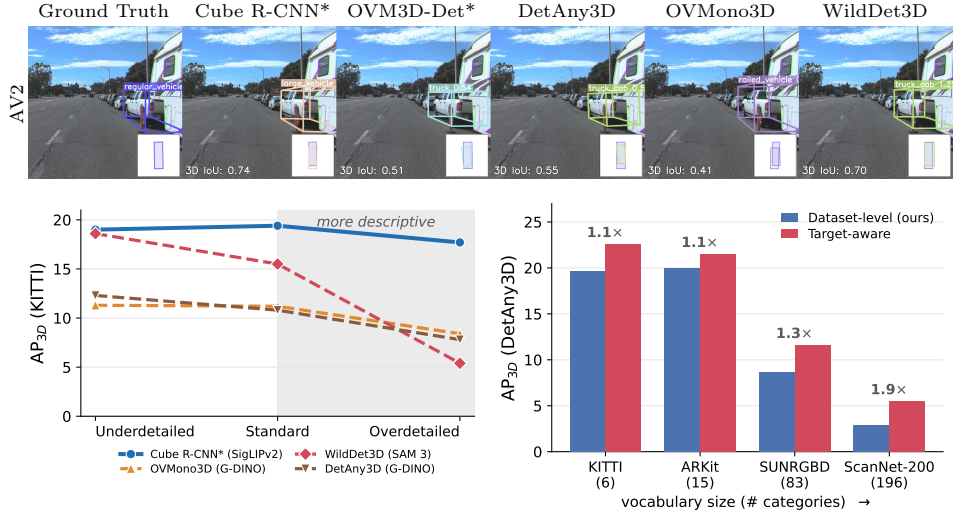

Open-vocabulary 3D object detection enables flexible 3D spatial reasoning through natural language queries and interactive user prompts. Such models have broad applications in autonomous driving \cite{khurana2024shelf, davidson2025refav, ovsep2024better, takmaz2025towards}, robot manipulation \cite{intelligence2025pi, wen2024foundationpose}, and extended reality (XR). However, unlike closed-world detectors that train and evaluate on a fixed set of pre-defined categories, open-vocabulary detectors must demonstrate strong zero-shot performance across diverse domains \cite{robicheaux2025roboflow100vl} to be practically useful.

\noindent Existing open-vocabulary methods report strong zero-shot results, yet each model measures success under a different protocol. Specifically, prior work partitions categories into disjoint {\tt base} and {\tt novel} sets, but we find that the split itself can (sometimes dramatically) inflate apparent generalization. More recently, target-aware metrics~\cite{zhang2025detany3d, yao2025ovmono3d} prompt each model with ground-truth categories in the test image. This protocol silently suppresses hallucinations and false positives, making it a poor proxy for real-world performance. Moreover, a deployed detector never knows which categories appear in a test image. Lastly, prior work only reports AP\textsubscript{3D}, which entangles localization quality with semantic correctness, so a low score cannot tell us where open-vocabulary models fail. Existing protocols therefore neither diagnose failures nor reflect realistic deployment conditions.

\noindent To address this gap, we introduce OV3D-Bench, a diagnostic benchmark that compares open-vocabulary monocular 3D detectors under deployment-realistic conditions. First, OV3D-Bench replaces the per-image target-aware oracle with dataset-level prompts. We make a deliberately weaker assumption than current benchmarks and prompt each model with the dataset's full vocabulary rather than each image's ground-truth classes, removing per-image privileged information while preserving deployment-time scene priors (\eg~urban scenes contain \texttt{cars} but not \texttt{sofas}). Importantly, this is not a return to closed-world evaluation, since the two protocols differ in when the label space is revealed; target class names are withheld at training time and are only exposed to the detector at test time. Since every class is scored on every image, established closed-set diagnostics, including per-class false positives, confusion matrices, and hallucination rates, can be evaluated for open-vocabulary models. Second, OV3D-Bench decouples the standard mAP\textsubscript{3D} metric along three axes: geometric accuracy (3D Class-Agnostic Recall), semantic robustness (confusion matrices and prompt sensitivity), and cross-domain generalization (mAP\textsubscript{3D} on unseen datasets). We report the three axes separately, which exposes strengths and weaknesses that a single metric hides. We evaluate seven representative detectors~\cite{lu2023ov3det, yang2024imov3d, brazil2023omni3d, huang2024ovm3ddet, zhang2025detany3d, yao2025ovmono3d, huang2026wilddet3d} and apply this protocol across seven indoor and outdoor datasets, including KITTI~\cite{geiger2012kitti}, nuScenes~\cite{caesar_nuscenes_2020}, Argoverse 2~\cite{Wilson2021Argoverse2} (AV2), SUNRGBD~\cite{song2015sunrgbd}, ScanNet-200~\cite{rozenberszki2022language}, ARKitScenes~\cite{Dehghan2021arkitscenes}, and Hypersim~\cite{Roberts2020Hypersim}. Third, we propose a frustratingly simple training-free baseline that remaps predictions from a frozen closed-vocabulary detector using a SigLIPv2~\cite{tschannen2025siglip2} encoder. Somewhat surprisingly, this baseline matches or surpasses purpose-built open-vocabulary methods~\cite{zhang2025detany3d} on unseen outdoor domains. This suggests that \emph{geometric localization is more mature, while open-vocabulary semantics remains the larger, unresolved gap}. Our breakdown analysis also shows that detectors routinely assign a semantically adjacent but incorrect label to a well-localized box. For example, five detectors correctly localize the same \texttt{regular vehicle} from Argoverse 2, but incorrectly identify it as a \texttt{truck}, \texttt{truck cab}, \texttt{railed vehicle}, or \texttt{large vehicle} (\cref{fig:teaser}). Furthermore, we find that open-vocabulary methods are sensitive to the level of detail used in prompt phrasing. We vary prompts from bare class names (\eg~\texttt{car}) to attribute-rich descriptions (\eg~\texttt{a detailed high-resolution photo of a car}), and find that rewording prompts from underdetailed to overdetailed hurts WildDet3D by $13.2$ AP\textsubscript{3D} ($18.6$ to $5.4$). In contrast, our remapping-based baseline applied to Cube R-CNN stays within $1.7$ AP\textsubscript{3D}. Lastly, we demonstrate that the target-aware oracle hides the cross-class false positives that matter most in safety-critical settings; DetAny3D scores $1.1\times$ higher under this protocol than under ours on ARKitScenes ($15$ categories), and $1.9\times$ higher on ScanNet-200 ($196$ categories).

We make three major contributions:
\begin{enumerate}
\item We propose an evaluation protocol that reports geometric, semantic, and cross-domain performance separately, rather than averaging them into a single AP\textsubscript{3D}. Furthermore, our protocol replaces the per-image oracle with dataset-level prompting to remove per-image privileged information while retaining scene-level priors.
\item We benchmark seven detectors and find that semantic vocabulary, not geometry, is the dominant failure mode. Detectors localize objects well but often classify the box incorrectly by assigning a semantically adjacent label.
\item We propose a simple yet effective training-free VLM remapping baseline that isolates the gap between localization quality and semantic accuracy.
\end{enumerate}

\section{Related Works}
\textbf{3D Detection Evaluation Practices.}
Omni3D~\cite{brazil2023omni3d} established the dominant protocol for 3D detection. It reports AP\textsubscript{3D} averaged across IoU thresholds and applies the same metric to both indoor and outdoor domains. Because AP\textsubscript{3D} scores recall, precision, and localization jointly, we cannot attribute a low score to a specific failure. Complementary diagnostics address parts of this problem. nuScenes~\cite{caesar_nuscenes_2020} reports true-positive metrics that isolate localization quality from detection, and long-tailed evaluation~\cite{peri2022towards,ma2025longtailed3ddetectionmultimodal} exposes hierarchical class confusions between common and rare categories. Both protocols target the closed-vocabulary setting, where the label space is fixed and semantic errors are bounded. In the open-vocabulary setting, the label space is itself the object of study, and no established protocol separates a mislocalized box from a misnamed one. We argue that open-vocabulary 3D detection has instead settled on evaluation practices that are optimistic rather than diagnostic. A common strategy partitions categories into {\tt base} and {\tt novel} sets, withholding {\tt novel} classes during training to simulate open-set generalization~\cite{lu2023ov3det, yang2024imov3d, huang2024ovm3ddet}. However, this protocol does not reflect realistic deployment, and the reported {\tt novel} AP depends heavily on which categories are withheld\cite{madan_revisiting_2024}. Results from papers adopting different splits are therefore not comparable. More recently, OVMono3D~\cite{yao2025ovmono3d} introduces (and DetAny3D~\cite{zhang2025detany3d} adopts) target-aware metrics that restrict evaluation to the categories present in each image. Because a detector is never prompted with a category absent from the image, it cannot produce a false positive for that category. Specifically, the cross-class errors that characterize open-vocabulary failures are removed rather than measured. We instead evaluate over an inclusive, dataset-level vocabulary and report geometry and semantics separately.

\begin{wraptable}{r}{0.46\textwidth}
\centering
\setlength{\tabcolsep}{3pt}
\renewcommand{\arraystretch}{0.9}
\resizebox{\linewidth}{!}{%
\begin{tabular}{>{\scriptsize}l|cccccc}
 & \rotatebox{90}{\scriptsize Open  Vocab.}
 & \rotatebox{90}{\scriptsize Mono. Input}
 & \rotatebox{90}{\scriptsize Text Prompt}
 & \rotatebox{90}{\scriptsize Box Prompt}
 & \rotatebox{90}{\scriptsize Point Prompt}
 & \rotatebox{90}{\scriptsize Large Scale} \\
\midrule
Cube R-CNN~\cite{brazil2023omni3d}   & \xmark & \cmark & \xmark & \xmark & \xmark & \xmark \\
OV-3Det~\cite{lu2023ov3det}          & \cmark & \xmark & \cmark & \xmark & \xmark & \xmark \\
ImOV3D~\cite{yang2024imov3d}         & \cmark & \xmark & \cmark & \xmark & \xmark & \xmark \\
OVM3D-Det~\cite{huang2024ovm3ddet}   & \xmark & \cmark & \xmark & \xmark & \xmark & \xmark \\
OVMono3D~\cite{yao2025ovmono3d}      & \cmark & \cmark & \cmark & \cmark & \xmark & \xmark \\
DetAny3D~\cite{zhang2025detany3d}    & \cmark & \cmark & \cmark & \cmark & \cmark & \cmark \\
WildDet3D~\cite{huang2026wilddet3d}  & \cmark & \cmark & \cmark & \cmark & \cmark & \cmark \\
\end{tabular}}
\caption{\textbf{Model Comparison.} We summarize the input modality, prompting modes, and training scale of the seven detectors in our benchmark above. A check mark denotes support.}
\label{tab:baselines}
\end{wraptable}

\noindent\textbf{VLMs for 3D Object Detection.}
Existing open-vocabulary 3D detectors leverage different strategies for aligning language with geometry (\cref{tab:baselines}). OV-3Det~\cite{lu2023ov3det} and ImOV3D~\cite{yang2024imov3d} classify learned point-cloud features against CLIP~\cite{radford2021clip} text embeddings using cosine similarity. OVM3D-Det~\cite{huang2024ovm3ddet} avoids explicit 3D supervision by lifting UniDepth and Grounded SAM~\cite{ren2024gsam} predictions into 3D pseudo-labels, though it still trains on fixed {\tt base} and {\tt novel} vocabularies from Omni3D~\cite{brazil2023omni3d}. OVMono3D~\cite{yao2025ovmono3d} embeds a frozen GroundingDINO~\cite{groundingDINO} in its training pipeline but evaluates on a predefined 50-category set from Omni3D~\cite{brazil2023omni3d}. DetAny3D~\cite{zhang2025detany3d} invokes GroundingDINO only at inference time. WildDet3D~\cite{huang2026wilddet3d} instead trains a promptable detector end-to-end on a large concept-annotated corpus. We take a deliberately minimal approach. Rather than proposing a new open-vocabulary architecture, we use SigLIPv2~\cite{tschannen2025siglip2} to remap a frozen closed-vocabulary detector's predictions into an open vocabulary without fine-tuning. This probe holds localization fixed and only varies the label space.

\section{OV3D-Bench: Towards A Better Open-Vocabulary 3D Benchmark}
Reporting a single overall AP\textsubscript{3D} overestimates open-vocabulary generalization and hides component failures. Instead, we introduce OV3D-Bench, an evaluation suite that decomposes performance along geometry, semantics, and domain-generalization.

\noindent\textbf{Geometric Accuracy.} We retain mAP\textsubscript{3D}~\cite{brazil2023omni3d} as our top-level detection metric for comparison with prior work, averaging AP over 3D IoU thresholds from $0.05$ to $0.5$ in steps of $0.05$, following Omni3D's protocol exactly. mAP\textsubscript{3D} conflates localization and category correctness, so a well-localized but mislabeled box scores as a failure. To isolate localization, we further report 3D Class-Agnostic Recall, which measures the fraction of ground-truth boxes matched to some prediction at IoU $\geq 0.15$, regardless of predicted category. We verified this choice against a confidence-ordered greedy assignment (as used by nuScenes~\cite{caesar_nuscenes_2020}), so our conclusions do not depend on the choice of assignment rule. Pairing mAP\textsubscript{3D} with 3D Class-Agnostic Recall separates semantic from localization error.

\noindent\textbf{Semantic Robustness.} Following prior long-tailed 3D detection analysis~\cite{peri2022towards,ma2025longtailed3ddetectionmultimodal}, we use \textit{confusion matrices} to dissect misclassifications. We first match each ground-truth 3D box to the highest IoU prediction under a one-to-one Hungarian assignment with IoU $\geq 0.5$, and discard unmatched boxes. Given $N$ classes, we then construct an $N \times N$ matrix where each cell $(i,j)$ reports the fraction of ground-truth class-$i$ objects that match to predictions of class $j$, normalized by per-class ground-truth counts. This isolates localization successes from semantic errors, \eg~{\tt shelves}$\to${\tt cabinet} (naming ambiguity) or {\tt fireplace}$\to${\tt television} (geometric similarity). Current target-aware protocols filter by observed categories and mask exactly these errors. We note that our confusion matrices and mAP\textsubscript{3D} treat all misclassifications equally; future work should incorporate semantic-distance weighting by penalizing semantically distant errors (\eg~{\tt toilet paper holder}$\to${\tt chair}) more than adjacent ones (\eg~{\tt sofa}$\to${\tt chair}) via the WordNet hierarchy or embedding distance.

\noindent Further, we argue that open-vocabulary detectors should be stable under natural prompt variations. We probe this with a \textit{template-based protocol}. For each category, we instantiate a family of prompts, following the prompt-type taxonomy of Lin et al.~\cite{lin2026userpromptingstrategiesprompt} and drawing within-group variations from Li et al.~\cite{li2025modelingvariantsprompts}. Specifically, we consider three prompt types by descriptiveness: \emph{underdetailed} (class names only, \eg~{\tt car}), \emph{standard} (CLIP-style~\cite{radford2021clip}, \eg~{\tt a photo of a car}), and \emph{overdetailed} (attribute-rich, \eg~{\tt a detailed high-resolution photo of a car}). We report mAP\textsubscript{3D} by prompt type, revealing the model's robustness to prompt variation. We also report the coefficient of variation of mAP\textsubscript{3D} across all templates ($\mathrm{CV}\,{=}\,\sigma/\mu$) as a scalar robustness summary. Lower is more robust.

\noindent\textbf{Domain Generalization.} To evaluate open-set generalization, we test on unseen datasets. Although the model may have seen some test classes during training (\eg train on a nuScenes {\tt truck} and test on an Argoverse 2 {\tt truck}), we argue that this cross-domain transfer represents more realistic deployment conditions than prior {\tt base}-{\tt novel} splits. We restrict \textit{cross-domain experiments} to same environment pairs (indoor $\leftrightarrow$ indoor, outdoor $\leftrightarrow$ outdoor), which reflects realistic semantic priors. OV-3Det~\cite{lu2023ov3det} and ImOV3D~\cite{yang2024imov3d} explored cross-domain transferability, but recent work has not adopted this practice. We extend this analysis to closed-set models using VLM-based remapping, described below.

\noindent\textbf{Training-Free VLM Remapping.} Isolating geometry from semantics requires a probe that holds localization fixed while varying only the vocabulary. Prior open-vocabulary detectors train new architectures that couple the two. Rather than training yet another architecture, we convert a frozen closed-vocabulary 3D detector into an open-vocabulary one through class remapping. At inference, we take an image, its ground-truth intrinsics, and a target category set. We use Omni3D's official taxonomy for the five in-domain datasets, and use the full native category list, taken as-is without de-duplication or merging, for the two zero-shot datasets (Argoverse 2 and ScanNet-200). This target set is fixed per dataset, independent of which categories occur in a given test image. We run the frozen detector to obtain candidate 3D boxes, project each box to 2D, and encode the crop with SigLIPv2~\cite{tschannen2025siglip2}. We then encode the dataset-level category names with the shared text encoder and remap each detection to its highest-similarity category above a threshold $\theta$, discarding it otherwise. Concretely, a detector trained only on \texttt{cars}, \texttt{pedestrians}, and \texttt{cyclists} can be evaluated on \texttt{trams} without retraining, since SigLIPv2 assigns a box resembling a tram to the \texttt{tram} class regardless of the detector's training categories. Holding localization fixed allows us to separate geometric errors from semantic errors. We denote SigLIPv2 remapping with $^*$ throughout (\eg~Cube R-CNN$^*$).

\section{Experiments}
\textbf{Datasets.}
Following Omni3D~\cite{brazil2023omni3d}, we evaluate on five in-domain datasets, namely KITTI~\cite{geiger2012kitti} (6 categories), nuScenes~\cite{caesar_nuscenes_2020} (9 categories), SUNRGBD~\cite{song2015sunrgbd} (83 categories), ARKitScenes~\cite{Dehghan2021arkitscenes} (15 categories), and Hypersim~\cite{Roberts2020Hypersim} (29 categories). For zero-shot evaluation we add  Argoverse 2~\cite{Wilson2021Argoverse2} (27 categories), with fine-grained classes like {\tt large vehicle}, {\tt regular vehicle}, and {\tt articulated bus} and  ScanNet-200~\cite{rozenberszki2022language} (196 categories), with fine-grained categories like {\tt hair dryer} and {\tt soap dispenser}. Such fine-grained categories are critical for stress testing open-vocabulary model performance.

\noindent\textbf{Baselines and Evaluation Details.}
We evaluate seven detectors, comprising five open-vocabulary models (OV-3Det~\cite{lu2023ov3det}, ImOV3D~\cite{yang2024imov3d}, DetAny3D~\cite{zhang2025detany3d}, OVMono3D~\cite{yao2025ovmono3d}, and WildDet3D~\cite{huang2026wilddet3d}) and two closed-vocabulary detectors (Cube R-CNN~\cite{brazil2023omni3d} and OVM3D-Det~\cite{huang2024ovm3ddet}). These models span diverse input modalities and training datasets (\cref{tab:baselines}). We describe the evaluation protocol below:

\begin{itemize}
    \item  \noindent\textit{Cube R-CNN~\cite{brazil2023omni3d}} is a closed-vocabulary monocular detector trained on Omni3D. We supply ground-truth intrinsics at inference for correct virtual-depth rescaling and apply our SigLIPv2 remapping with threshold $\theta=0.05$.

    \item  \noindent\textit{OVM3D-Det~\cite{huang2024ovm3ddet}} trains without 3D-label supervision using metric depth~\cite{piccinelli2024unidepth}. Despite its open-vocabulary claim, the authors release dataset-specific models trained on fixed {\tt base} and {\tt novel} vocabularies from Omni3D~\cite{brazil2023omni3d}. We therefore treat it as closed-vocabulary and follow the same protocol as Cube R-CNN.

    \item  \noindent\textit{OV-3Det~\cite{lu2023ov3det} and ImOV3D~\cite{yang2024imov3d}} are indoor point-cloud-based open-vocabulary detectors. To comply with the monocular setting, we build point clouds from UniDepth~\cite{piccinelli2024unidepth} predictions with ground-truth intrinsics. During inference, we align text and point cloud features with CLIP~\cite{radford2021clip}.

    \item  \noindent\textit{OVMono3D~\cite{yao2025ovmono3d}} is an open-vocabulary monocular detector trained on Omni3D~\cite{brazil2023omni3d}. We use OVMono3D-LIFT, which lifts GroundingDINO~\cite{groundingDINO} 2D detections to 3D.

    \item  \noindent\textit{DetAny3D~\cite{zhang2025detany3d}} is a promptable foundation model that generalizes across cameras, trained on a large corpus (Omni3D, Argoverse 2, and ScanNet). At inference, we feed GroundingDINO~\cite{groundingDINO} 2D boxes to its 3D head, following OVMono3D's protocol, we use a low box threshold ($0.001$) for high recall, with each detection's category resolved by matching GroundingDINO's per-token caption logits in token space (text threshold $0.25$) rather than its decoded phrase string. To maintain consistency with our other baselines, we provide ground truth intrinsics.

    \item  \noindent\textit{WildDet3D~\cite{huang2026wilddet3d}} unifies SAM3's~\cite{carion2025sam3} promptable detection with DINOv2's geometry, avoiding external detectors like GroundingDINO. It trains on the largest 3D detection corpus to date (1M images and 13k categories, $138\times$ Omni3D). Given its high recall, we apply NMS with an IoU threshold of $0.3$ and a confidence threshold of $0.8$.
\end{itemize}

\begin{table*}[ht!]
\centering
\small
\setlength{\tabcolsep}{4pt}
\resizebox{\textwidth}{!}{%
\begin{tabular}{l|ccccccc}
& SUNRGBD & ARKitScenes & Hypersim & ScanNet-200 & KITTI & nuScenes & Argoverse 2 \\
\hline
Cube R-CNN*
& \cellcolor{green!25}0.523 & \cellcolor{green!25}0.335 & \cellcolor{green!25}0.085 & \cellcolor{yellow!25}0.157
& \cellcolor{green!25}0.716 & \cellcolor{green!25}0.389 & \cellcolor{yellow!25}0.286 \\
\hline
OVMono3D
& \cellcolor{green!25}0.586 & \cellcolor{green!25}\textbf{0.389} & \cellcolor{green!25}\textbf{0.129} & \cellcolor{yellow!25}0.218
& \cellcolor{green!25}\textbf{0.729} & \cellcolor{green!25}\textbf{0.396} & \cellcolor{yellow!25}0.279 \\
DetAny3D
& \cellcolor{green!25}0.440 & \cellcolor{green!25}0.171 & \cellcolor{green!25}0.064 & \cellcolor{green!25}0.146
& \cellcolor{green!25}0.642 & \cellcolor{green!25}0.281 & \cellcolor{green!25}0.142 \\
WildDet3D
& \cellcolor{green!25}\textbf{0.640} & \cellcolor{green!25}0.323 & \cellcolor{green!25}0.117 & \cellcolor{yellow!25}\textbf{0.389}
& \cellcolor{green!25}0.597 & \cellcolor{green!25}0.304 & \cellcolor{yellow!25}\textbf{0.326} \\
\end{tabular}%
}
\caption{\textbf{3D Class-Agnostic Recall ($\uparrow$).} No method dominates every dataset; recall varies far more \emph{across datasets} (dense indoor scenes like Hypersim and ScanNet-200 are hardest) than across methods within a dataset. Green marks in-domain datasets (Omni3D); yellow marks unseen ones (green for DetAny3D, whose training corpus includes Argoverse 2 and ScanNet). We mark the best method per dataset in \textbf{bold}.
}
\label{tab:tp_iou3d}
\vspace{-1em}
\end{table*}
\begin{table*}[ht!]
\centering
\small
\setlength{\tabcolsep}{4pt}
\resizebox{\textwidth}{!}{%
\begin{tabular}{l|ccccccc}
& SUNRGBD & ARKitScenes & Hypersim & ScanNet-200 & KITTI & nuScenes & Argoverse 2 \\
\hline
Cube R-CNN*
& \cellcolor{green!25}6.38 & \cellcolor{green!25}21.33 & \cellcolor{green!25}5.33 & \cellcolor{yellow!25}2.22
& \cellcolor{green!25}17.64 & \cellcolor{green!25}15.49 & \cellcolor{yellow!25}\textbf{11.10} \\
\hline
OVMono3D
& \cellcolor{green!25}7.18 & \cellcolor{green!25}\textbf{28.31} & \cellcolor{green!25}\textbf{6.95} & \cellcolor{yellow!25}2.44
& \cellcolor{green!25}15.96 & \cellcolor{green!25}13.81 & \cellcolor{yellow!25}2.26 \\
DetAny3D
& \cellcolor{green!25}\textbf{8.81} & \cellcolor{green!25}19.92 & \cellcolor{green!25}6.23 & \cellcolor{green!25}\textbf{2.89}
& \cellcolor{green!25}19.65 & \cellcolor{green!25}12.83 & \cellcolor{green!25}0.82 \\
WildDet3D
& \cellcolor{green!25}5.92 & \cellcolor{green!25}23.24 & \cellcolor{green!25}6.43 & \cellcolor{yellow!25}2.23
& \cellcolor{green!25}\textbf{19.92} & \cellcolor{green!25}\textbf{18.00} & \cellcolor{yellow!25}4.82 \\
\end{tabular}%
}
\caption{\textbf{Standard mAP$_\text{3D}$ Detection Evaluation ($\uparrow$).} No method wins outright, with each detector leading on at most two of seven datasets. Green marks in-domain datasets (Omni3D); yellow marks unseen ones. Argoverse 2 and ScanNet are in the DetAny3D training corpus, so those results are not zero-shot. We mark the best method per dataset in \textbf{bold}.
}
\label{tab:standard_metrics}
\vspace{-1em}
\end{table*}

\begin{table*}[ht!]
\centering
\small
\setlength{\tabcolsep}{4pt}
\begin{subtable}[t]{0.5\linewidth}
\centering
\resizebox{\linewidth}{!}{%
\begin{tabular}{ll|cccc}
Method & Train & SUNRGBD & ARKit & Hypersim & ScanNet-200 \\
\hline
\multirow{2}{*}{OVM3D-Det*}
 & SU & \cellcolor{green!25}\textbf{0.583} & \cellcolor{yellow!25}0.273 & \cellcolor{yellow!25}\textbf{0.080} & \cellcolor{yellow!25}0.179 \\
 & AR & \cellcolor{yellow!25}0.450 & \cellcolor{green!25}\textbf{0.298} & \cellcolor{yellow!25}0.062 & \cellcolor{yellow!25}0.156 \\
\hline
\multirow{2}{*}{OV-3Det}
 & SU & \cellcolor{green!25}0.403 & \cellcolor{yellow!25}0.237 & \cellcolor{yellow!25}0.046 & \cellcolor{yellow!25}0.163 \\
 & SC & \cellcolor{yellow!25}0.350 & \cellcolor{yellow!25}0.180 & \cellcolor{yellow!25}0.028 & \cellcolor{green!25}0.146 \\
\hline
\multirow{2}{*}{ImOV3D}
 & SU & \cellcolor{green!25}0.490 & \cellcolor{yellow!25}0.235 & \cellcolor{yellow!25}0.059 & \cellcolor{yellow!25}\textbf{0.186} \\
 & SC & \cellcolor{yellow!25}0.449 & \cellcolor{yellow!25}0.225 & \cellcolor{yellow!25}0.060 & \cellcolor{green!25}0.175 \\
\end{tabular}}
\caption{3D Class-Agnostic Recall.}
\label{tab:tp_iou3d_ds}
\end{subtable}\hfill
\begin{subtable}[t]{0.48\linewidth}
\centering
\resizebox{\linewidth}{!}{%
\begin{tabular}{ll|cccc}
Method & Train & SUNRGBD & ARKit & Hypersim & ScanNet-200 \\
\hline
\multirow{2}{*}{OVM3D-Det*}
 & SU & \cellcolor{green!25}\textbf{5.16} & \cellcolor{yellow!25}3.02 & \cellcolor{yellow!25}1.44 & \cellcolor{yellow!25}1.39 \\
 & AR & \cellcolor{yellow!25}4.55 & \cellcolor{green!25}\textbf{16.56} & \cellcolor{yellow!25}\textbf{2.94} & \cellcolor{yellow!25}\textbf{2.03} \\
\hline
\multirow{2}{*}{OV-3Det}
 & SU & \cellcolor{green!25}0.69 & \cellcolor{yellow!25}1.10 & \cellcolor{yellow!25}0.66 & \cellcolor{yellow!25}0.13 \\
 & SC & \cellcolor{yellow!25}0.45 & \cellcolor{yellow!25}0.95 & \cellcolor{yellow!25}0.21 & \cellcolor{green!25}0.17 \\
\hline
\multirow{2}{*}{ImOV3D}
 & SU & \cellcolor{green!25}1.57 & \cellcolor{yellow!25}3.08 & \cellcolor{yellow!25}1.17 & \cellcolor{yellow!25}0.40 \\
 & SC & \cellcolor{yellow!25}1.04 & \cellcolor{yellow!25}2.03 & \cellcolor{yellow!25}0.94 & \cellcolor{green!25}0.27 \\
\end{tabular}}
\caption{Standard mAP$_\text{3D}$.}
\label{tab:standard_ds}
\end{subtable}
\caption{\textbf{Cross-Dataset Transfer Costs Recognition, Not Geometry.} We train a separate model per dataset and evaluate each in-domain and on out-of-domain transfer, across SUNRGBD (SU), ScanNet-200 (SC), and ARKitScenes (AR) on (\subref{tab:tp_iou3d_ds}) 3D Class-Agnostic Recall and (\subref{tab:standard_ds}) standard mAP\textsubscript{3D}. Recall shifts modestly between in- and out-of-domain models, while mAP\textsubscript{3D} favors the in-domain model by as much as $5.5\times$. Green marks in-domain results; yellow marks cross-domain transfer. Higher values are better for both metrics, and we mark the best setting per dataset in \textbf{bold}.}
\vspace{-1em}
\end{table*}

\begin{table}[t]
\setlength{\tabcolsep}{6pt}
\centering
\resizebox{0.6\columnwidth}{!}{%
\begin{tabular}{l|ccc}
Train & KITTI & nuScenes & Argoverse 2 \\
\hline
KITTI & \cellcolor{green!25}\textbf{12.87} & \cellcolor{yellow!25}2.78 & \cellcolor{yellow!25}1.37 \\
nuScenes & \cellcolor{yellow!25}9.69 & \cellcolor{green!25}\textbf{7.95} & \cellcolor{yellow!25}\textbf{14.99} \\
\end{tabular}%
}
\caption{\textbf{Only nuScenes Pre-Training Transfers Outdoors.} We train OVM3D-Det* on KITTI and on nuScenes and evaluate both on all three outdoor datasets. The nuScenes-trained model recovers $75\%$ of in-domain KITTI mAP\textsubscript{3D}, while the KITTI-trained model recovers only $35\%$ on nuScenes. Green indicates in-domain; yellow indicates cross-domain transfer; \textbf{bold} denotes the best configuration.}
\label{tab:ovm3d_outdoor}
\end{table}

\begin{figure*}[t!]
    \centering
    \includegraphics[trim={0 0 1.5em 0}, clip, width=\linewidth]{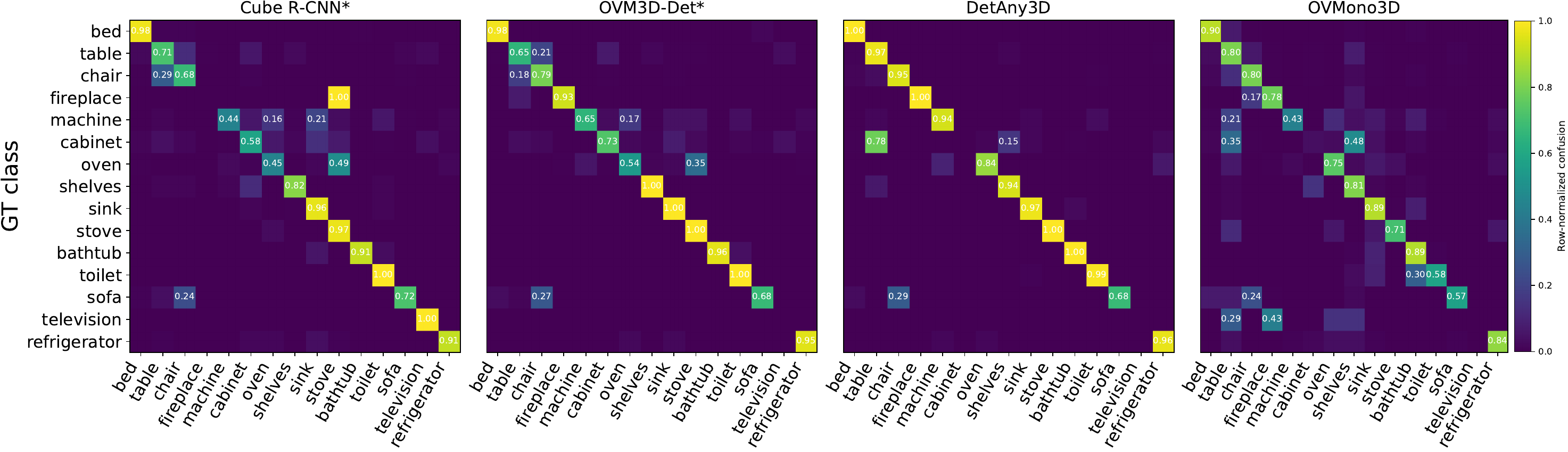}
    \caption{\textbf{Detectors Share the Same Confusions.} We visualize confusion matrices over the $15$ ARKitScenes categories for Cube R-CNN*, OVM3D-Det*, DetAny3D, and OVMono3D. Each row is normalized by its ground-truth count over matched boxes, and the diagonal gives per-class recall. Furniture, appliances and storage absorb nearly all off-diagonal probability mass across all four detectors, while {\tt sink} and {\tt bathtub} stay cleanly separated. Best viewed in color and zoomed in.}
    \label{fig:confmat_arkit}
\end{figure*}
\begin{figure*}[!t]
    \centering
    \begin{overpic}[width=\linewidth,]{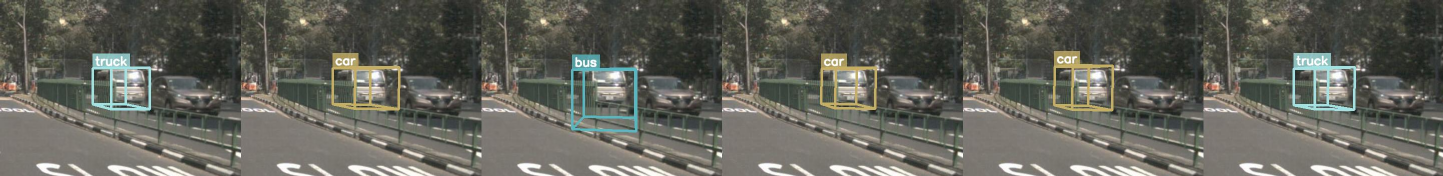}
        \put(1,12.9){\scriptsize Ground Truth}
        \put(18,12.9){\scriptsize Cube R-CNN*}
        \put(34.5,12.9){\scriptsize OVM3D-Det*}
        \put(53,12.9){\scriptsize DetAny3D}
        \put(69,12.9){\scriptsize OVMono3D}
        \put(85,12.9){\scriptsize WildDet3D}
        \put(-2,2.5){\rotatebox{90}{\scriptsize nuScenes}}
    \end{overpic}\\[0.15em]
    \begin{overpic}[width=\linewidth]{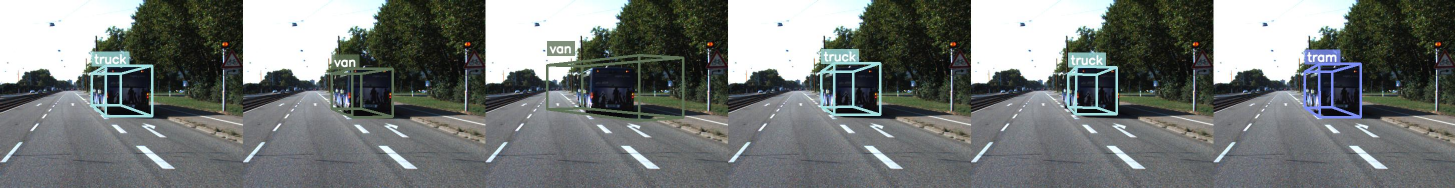}
        \put(-2,3){\rotatebox{90}{\scriptsize KITTI}}
    \end{overpic}\\[0.15em]
    \begin{overpic}[width=\linewidth]{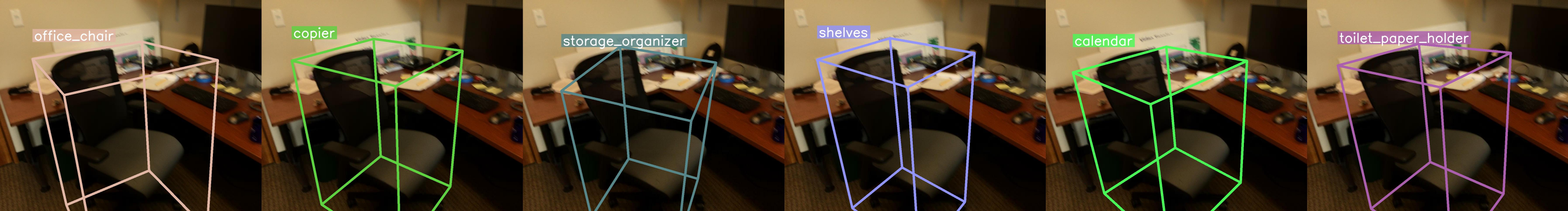}
        \put(-2.0,3.3){\rotatebox{90}{\scriptsize ScanNet}}
    \end{overpic}\\[0.15em]
    \begin{overpic}[width=\linewidth]{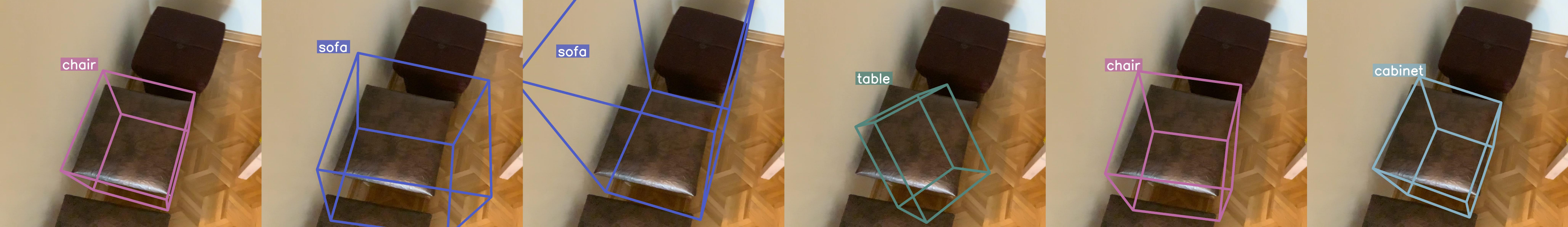}
        \put(-1.8,0.5){\rotatebox{90}{\scriptsize ARKitScenes}}
    \end{overpic}
    \caption{\textbf{Failures Are Semantic, Not Geometric.} Ground truth box alongside predictions from each of the five detectors. Boxes are well aligned to the object across nearly every detector, while the predicted labels disagree.}
    \label{fig:qual}
\end{figure*}

\noindent\textbf{3D Class-Agnostic Recall Evaluation.}
We report the ratio of ground-truth boxes matched by some prediction at IoU $\geq 0.15$, under a class-agnostic assignment that credits a well-localized box regardless of its predicted category (\cref{tab:tp_iou3d}). Recall varies far more \emph{across datasets} than across methods. While the values are low on Hypersim and ScanNet-200 due to dense indoor scenes with many small, occluded objects, recall reaches up to $0.73$ on the sparser outdoor KITTI. No single method dominates every dataset: OVMono3D leads on four of seven (ARKitScenes, Hypersim, KITTI, nuScenes) and WildDet3D on the remaining three (SUNRGBD, ScanNet-200, Argoverse 2), while Cube R-CNN* and DetAny3D trail throughout.

\noindent\textbf{3D Detection Confusion Matrices.}
OVMono3D over-predicts {\tt tables} and {\tt shelves}, whereas DetAny3D favors {\tt chairs}, which are prevalent in indoor training datasets (\cref{fig:confmat_arkit}). The remapped Cube R-CNN$^*$ and OVM3D-Det$^*$ confuse spatially co-located {\tt tables} and {\tt chairs}, due to projective ambiguity in the image crops passed to SigLIPv2. Text-encoder ambiguities (\eg~{\tt armchair}$\rightarrow$\ {\tt sofa chair}) compound these errors. Errors concentrate along a small number of categories; furniture ({\tt bed}, {\tt table}, {\tt chair}, {\tt sofa}) and appliances and storage ({\tt machine}, {\tt cabinet}, {\tt oven}, {\tt shelves}) absorb nearly all off-diagonal probability mass, while {\tt sink} and {\tt bathtub} are cleanly separated by all four detectors, suggesting that the residual errors follow the ARKitScenes label space rather than any one detector.
\begin{wrapfigure}{r}{0.4\textwidth}
    \centering
    \includegraphics[width=\linewidth]{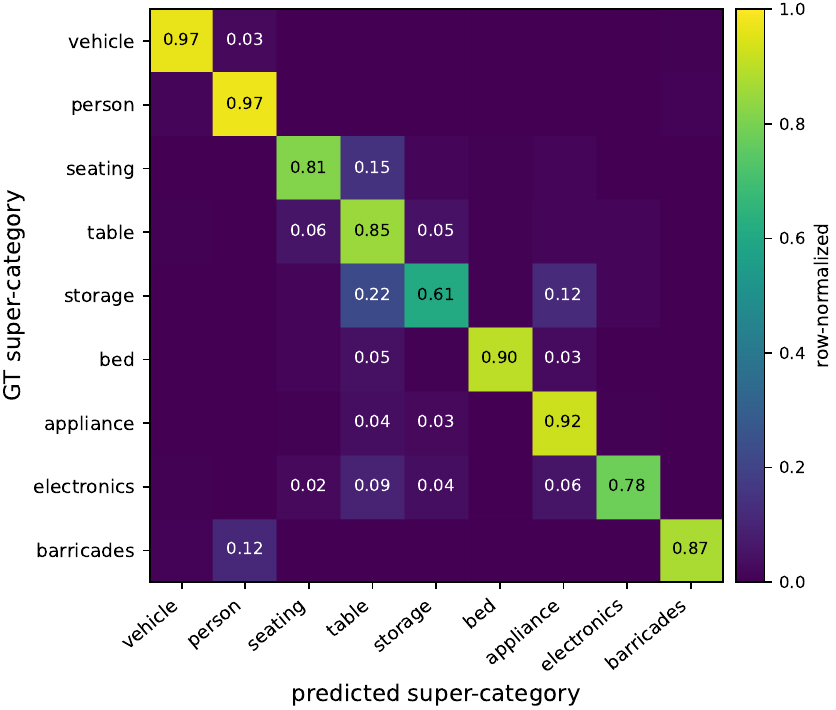}
    \caption{\textbf{Confusion Concentrates Among Indoor Furniture.} Predictions aggregated over all datasets and detectors, row-normalized by GT super-category.}
    \label{fig:supercat}
\end{wrapfigure}

\noindent We project all $267$ classes in OV3D-Bench onto a shared coarse ontology, which gives a vocabulary-agnostic view of model confusion (\cref{fig:supercat}). We manually assign each fine-grained class to one of $9$ super-categories, namely \texttt{vehicle}, \texttt{person}, \texttt{seating}, \texttt{table}, \texttt{storage}, \texttt{bed}, \texttt{appliance}, \texttt{electronics}, and \texttt{barricades}, by matching category-name substrings (\eg~\texttt{cabinet}, \texttt{shelves}, and \texttt{wardrobe} all fall under \texttt{storage}). We find that only $15\%$ of confusions cross super-category boundaries. The diagonal ranges from $0.61$ (\texttt{storage}) to $0.97$ (\texttt{vehicle}, \texttt{person}). Interestingly, confusion is near-zero for vehicles and $5$--$22\%$ among co-located indoor furniture, with one exception (\texttt{barricades}$\rightarrow$\texttt{person}, $12\%$). Notably, \texttt{table} absorbs off-diagonal mass from five of the eight other categories, including the two largest errors in the matrix (\texttt{storage}$\to$\texttt{table} at $0.22$ and \texttt{seating}$\to$\texttt{table} at $0.15$), while emitting at most $0.06$ itself. This suggests that detectors recover coarse object identity, but not fine-grained labels. \Cref{fig:qual} shows representative cases, where boxes are well aligned to the object across nearly every detector while the labels disagree. Outdoor errors are taxonomy-boundary disagreements (\eg~the AV2 example in \cref{fig:teaser}: \texttt{truck}, \texttt{truck cab}, \texttt{railed vehicle}, and \texttt{large vehicle} for one \texttt{regular vehicle}), whereas indoor errors span categories with no shared geometry, indicating that the residual error is in mapping onto a dataset's vocabulary, not in recovering an object's spatial extent.

\begin{table}[t]
\begin{minipage}[t]{0.48\linewidth}
\centering
\small
\setlength{\tabcolsep}{3pt}
\resizebox{\linewidth}{!}{%
\begin{tabular}{l|c|cc|cc}
& & \multicolumn{2}{c|}{DetAny3D} & \multicolumn{2}{c}{OVMono3D} \\
Dataset & \#CLS & DLP & TAP & DLP & TAP \\
\midrule
KITTI       & 6   & 19.65 & 22.55 & 15.96 & 18.56 \\
ARKitScenes & 15  & 19.92 & 21.47 & 28.31 & 30.57 \\
SUNRGBD     & 83  & 8.70  & 11.59 & 7.18  & 10.23 \\
ScanNet-200 & 196 & 2.85  & 5.47  & 2.44  & 5.29  \\
\end{tabular}%
}
\caption{\textbf{Target-Aware Prompting Inflates mAP\textsubscript{3D}, Most
Where Scores Are Lowest.} We evaluate identical predictions under dataset-level (DLP) and target-aware (TAP) prompting for DetAny3D and OVMono3D across datasets of increasing vocabulary size. TAP raises AP by a similar absolute margin in every setting, so its relative inflation is largest exactly where the DLP score is smallest.}
\label{tab:target_aware}
\end{minipage}\hfill
\begin{minipage}[t]{0.48\linewidth}
\centering
\small
\setlength{\tabcolsep}{3pt}
\resizebox{\linewidth}{!}{%
\begin{tabular}{l|cccccc}
Category & car & ped. & van & cyc. & truck & tram \\
\#GT     & 14232 & 2445 & 1586 & 892 & 585 & 273 \\
\midrule
DLP & 42.0 & 15.3 & 14.3 & 10.5 & 26.7 & 9.1 \\
TAP  & 42.1 & 16.0 & 16.6 & 11.5 & 37.4 & 11.7 \\
$\Delta$      & +0.1 & +0.7 & +2.4 & +0.9 & +10.7 & +2.6 \\
\end{tabular}%
}
\caption{\textbf{Target-Aware Gains Concentrate on Cross-Class Confusions.} We compare per-category AP\textsubscript{3D} for DetAny3D on KITTI under both protocols. \texttt{car} performance is unchanged, while \texttt{truck}, \texttt{van}, and \texttt{tram}, the classes most often confused with \texttt{car}, gain the most.}
\label{tab:kitti_targetaware}
\end{minipage}
\vspace{-1em}
\end{table}

\noindent\textbf{Dataset-Level mAP\textsubscript{3D} Evaluation.}
mAP\textsubscript{3D}~\cite{brazil2023omni3d} jointly scores recall, precision, and localization over \emph{all} predictions, surfacing the semantic gap that target-aware protocols hide. Each detector in \cref{tab:standard_metrics} leads on at most two of seven datasets, and the closed-set Cube R-CNN* and OVM3D-Det* match open-vocabulary OVMono3D and DetAny3D (\cref{tab:standard_metrics}, with the dataset-level protocol in \cref{tab:standard_ds}). We find that nuScenes pre-training drives large Argoverse gains. Cube R-CNN* reaches $11.10$ mAP, $+390\%$ over OVMono3D's $2.26$ and $2.3\times$ the next-best method, and nuScenes-trained OVM3D-Det* reaches $14.99$, far exceeding DetAny3D's $0.82$ (\cref{tab:ovm3d_outdoor}). However, this transfer is asymmetric. The nuScenes-trained model recovers $75\%$ of in-domain KITTI mAP\textsubscript{3D}, while the KITTI-trained model recovers only $35\%$ on nuScenes and collapses to $1.37$ on Argoverse 2. Notably, the nuScenes-trained model scores higher on Argoverse 2 than in domain, which we posit can be attributed to nuScenes averaging over hard classes near zero. Cube R-CNN* nonetheless trails both GroundingDINO baselines on all four indoor datasets, by up to $2.43$ mAP\textsubscript{3D} on SUNRGBD, suggesting that the outdoor label space is small enough for remapping to cover, unlike cluttered indoor vocabularies. 

\noindent Comparing in- against out-of-domain models per dataset (\cref{tab:tp_iou3d_ds,tab:standard_ds}), mAP\textsubscript{3D} favors the in-domain model on five of six pairs, by up to $5.5\times$ on ARKitScenes, while 3D Class-Agnostic Recall shifts far more modestly across the same pairs (e.g.\ OVM3D-Det* on SUNRGBD: $0.583\!\to\!0.450$): geometry transfers across domains, recognition does not. Out-of-domain transfer degrades primarily due to hallucinated classes and false positives, and ScanNet-trained point-cloud models generalize poorly to ARKitScenes and Hypersim. 

\noindent Comparing to the class-agnostic evaluation in \cref{tab:tp_iou3d}, recall values exceed the corresponding mAP\textsubscript{3D} scores (as a fraction) on every dataset we test, which validates the premise that detectors consistently localize more objects than mAP\textsubscript{3D} credits them for. Since class-agnostic recall ignores the predicted category entirely, this gap cannot be a localization failure, but the cost of getting the category (and confidence ranking) wrong. \textit{Geometric localization is therefore more mature, while semantic recognition remains the larger, unresolved gap.}

\begin{figure*}[t]
\centering
\begin{overpic}[width=\linewidth]{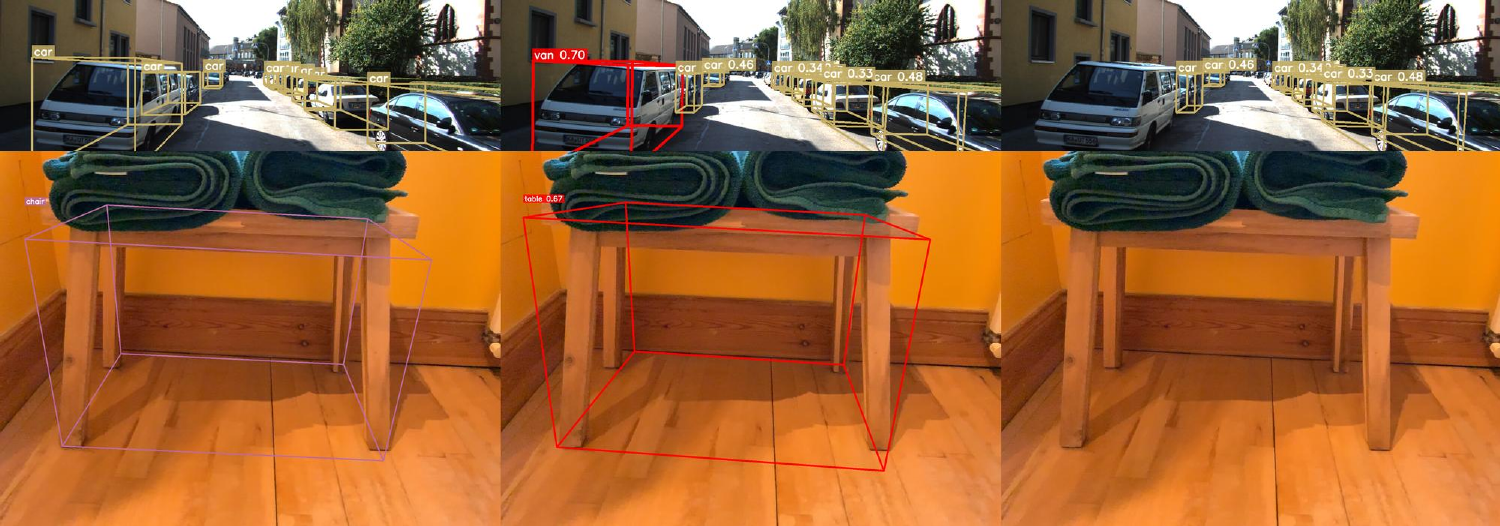}
    \put(11,35.8){\scriptsize Ground Truth}
    \put(39,35.8){\scriptsize Dataset-Level Protocol}
    \put(73,35.8){\scriptsize Target-Aware Protocol}
    \put(-2.3,27.0){\rotatebox{90}{\scriptsize KITTI}}
    \put(-2.3,6.5){\rotatebox{90}{\scriptsize ARKitScenes}}
    \put(87,33){\colorbox{white}{\textcolor{red}{\scriptsize $+2.4$\,AP}}}
    \put(87,21.5){\colorbox{white}{\textcolor{red}{\scriptsize $+2.0$\,AP}}}
\end{overpic}
\caption{\textbf{The Target-Aware Protocol Inflates AP by Discarding Hallucinated
Classes.} We visualize DetAny3D predictions on KITTI (top) and ARKitScenes (bottom) images to compare the target-aware (TAP) and dataset-level (DLP) protocols. The TAP oracle deletes hallucinated boxes for categories that do not occur in the image rather than penalizing them, raising AP by $2.4$ on KITTI and $2.0$ on ARKitScenes. We mark detections of categories absent from that image's ground truth in \textcolor{red}{red}.}
\label{fig:protocol}
\end{figure*}
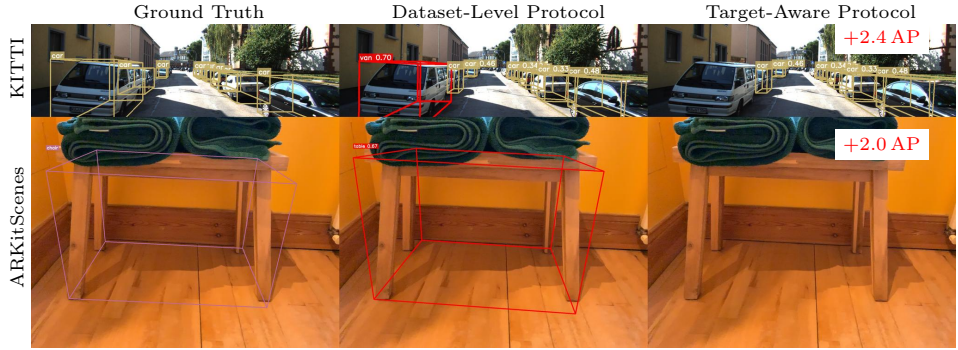

\noindent\textbf{Comparison with the Target-Aware Protocol.}
Recent detectors report under a \emph{target-aware} protocol~\cite{yao2025ovmono3d,zhang2025detany3d} that prompts each image only with its ground-truth categories. We re-score the same predictions under both protocols (\cref{tab:target_aware}), and find that the oracle inflates mAP\textsubscript{3D} for every dataset and method by between $+8\%$ and $+117\%$ relative, nearly doubling DetAny3D's large-vocabulary ScanNet-200 performance and more than doubling OVMono3D's (DetAny3D $2.85\!\to\!5.47$, OVMono3D $2.44\!\to\!5.29$). KITTI shows this mechanism in practice (\cref{tab:kitti_targetaware}). For example, \texttt{car} class performance is unchanged, while performance for classes most commonly confused with \texttt{car} is strongly inflated (e.g., \texttt{truck} $26.7\!\to\!37.4$, $1.4\times$; \texttt{van} $+2.4$; \texttt{tram} $+2.6$). The detector confidently mislabels cars as \texttt{van} and chairs as \texttt{table}, and the oracle deletes these false positives wherever those classes are absent (\cref{fig:protocol}). Target-aware evaluation thus conceals the errors most consequential for safe deployment, while our dataset-level protocol exposes them.

\begin{table}[t]
\centering
\scriptsize
\setlength{\tabcolsep}{2.5pt}
\begin{tabular}{lll|ccc|ccc|c}
\multirow{2}{*}{Encoder} & \multirow{2}{*}{Prompt usage} & \multirow{2}{*}{Tokenizer} & \multicolumn{3}{c|}{ARKitScenes} & \multicolumn{3}{c|}{KITTI} & CV\% \\
 & & & Und. & Std. & Ovr. & Und. & Std. & Ovr. & $\downarrow$ \\
\midrule
SigLIPv2\textsuperscript{*} & Remapping & SentencePiece & 26.8 & \textbf{27.7} & 23.3 & 19.0 & \textbf{19.4} & 17.7 & \textbf{8} \\
CLIP\textsuperscript{*}     & Remapping & BPE           & 19.1 & 21.0 & 19.5 & 14.8 & 10.7 & 12.4 & 12 \\
\midrule
OVMono3D~\cite{yao2025ovmono3d}     & Grounding & BERT & 26.4 & 22.2 & 21.3 & 11.3 & 11.2 & 8.4 & 20 \\
WildDet3D~\cite{huang2026wilddet3d} & Prompt Cond. & BPE & 28.9 & 33.1 & 21.7 & 18.6 & 15.5 & 5.4 & 34 \\
DetAny3D~\cite{zhang2025detany3d}   & Grounding & BERT & 16.6 & 14.3 & 13.8 & 12.3 & 10.8 & 7.8 & 28 \\
\end{tabular}
\caption{\textbf{Prompt Robustness.} We evaluate mAP\textsubscript{3D} on a fixed
500-image subset, grouped into \emph{underdetailed} (Und.),
\emph{standard} (Std.), and \emph{overdetailed} (Ovr.) templates (5 variations per template).
Encoders that only remap fixed detections vary least across prompt phrasings, while those that condition detections on the prompt vary far more.
\textsuperscript{*} marks the encoder used to remap Cube R-CNN detections (SigLIPv2 or CLIP). ``Grounding''
denotes GroundingDINO, and ``Prompt Cond.'' denotes SAM 3-style
prompt-conditioned detection. CV\% is the coefficient of variation across all
15 templates, averaged over both datasets ($\downarrow$ is better).}
\vspace{-1em}
\label{tab:sensitivity}
\end{table}

\noindent \textbf{Prompt Sensitivity.}
We evaluate five encoders in \cref{tab:sensitivity}. Our remapping strategy uses the SigLIPv2 and CLIP contrastive encoders, OVMono3D and DetAny3D use the GroundingDINO~\cite{groundingDINO} grounding encoder, and WildDet3D uses a jointly-trained SAM3 encoder~\cite{huang2026wilddet3d}. This is a controlled probe rather than the main benchmark protocol. Specifically, we run each method on a fixed $500$-image subset across the same $15$ templates at high recall, so only relative quantities are meaningful, namely the coefficient of variation and the degradation across prompt types. These values are not comparable to \cref{tab:standard_metrics}, where we prompt each detector with dataset category names on the full test sets.

\noindent The \emph{standard} group here denotes CLIP-style templates (\eg~\texttt{a photo of a \{name\}}), not the default prompting of \cref{tab:standard_metrics}. DetAny3D can therefore score well in \cref{tab:standard_metrics} while degrading sharply once we impose a common template. Our remapping sidesteps this failure mode, making it the most prompt-stable.

\noindent We find that \emph{standard} prompts are near-optimal for the remapping encoders, consistent with the 2D finding of Lin et al.~\cite{lin2026userpromptingstrategiesprompt}, though CLIP on KITTI peaks on class names alone. This does not generalize to detectors that condition on the prompt, since OVMono3D and DetAny3D both peak on \emph{underdetailed} prompts. Importantly, \emph{overdetailed} phrasing hurts every encoder, most severely for WildDet3D, which drops $11.4$ mAP\textsubscript{3D} on ARKitScenes and $10.1$ on KITTI relative to \emph{standard}, against at most $4.4$ for either remapping encoder.

\noindent Notably, the tokenizer does not directly impact performance, since WildDet3D and our CLIP remapping share a BPE tokenizer yet differ ${\sim}3\times$ in CV. Instead, \emph{how the prompt is consumed} matters most. Remapping encoders rank categories by cosine similarity, so descriptive tokens perturb the embedding without upsetting the ranking ($8\%$ CV for SigLIPv2), whereas prompt-conditioned detectors, trained on short noun phrases, suppress detections under descriptive phrasing. Neither large-scale training data nor contrastive pretraining confers robustness once detectors are directly conditioned on the prompt.

\begin{table}[t]
\centering
\small
\setlength{\tabcolsep}{4pt}
\resizebox{\columnwidth}{!}{%
\begin{tabular}{ll|ccc|cc|c}
\multirow{2}{*}{Method} & \multirow{2}{*}{Dataset} & \# CLS & Full & Random & \multicolumn{2}{c|}{$10$ Class Split} & Fav./ \\
 & & (GT) & Vocab. & ($\mu\pm\sigma$) & Adverse & Favorable & Full \\
\midrule
\multirow{2}{*}{DetAny3D}  & SUNRGBD     & 82  & 8.81 & 8.7$\pm$3.9 & 0.0 & 38.9 & 4.4$\times$ \\
                           & ScanNet-200 & 174 & 2.89 & 2.9$\pm$1.7 & 0.0 & 22.2 & 7.7$\times$ \\
\multirow{2}{*}{OVMono3D}  & SUNRGBD     & 82  & 7.18 & 7.2$\pm$3.9 & 0.0 & 37.9 & 5.3$\times$ \\
                           & ScanNet-200 & 174 & 2.44 & 2.4$\pm$1.7 & 0.0 & 21.2 & 8.7$\times$ \\
\end{tabular}}
\caption{\textbf{Vocabulary Subset Selection Outweighs the Choice of Detector.} For fixed predictions (model unchanged, having seen all classes) we report
mAP\textsubscript{3D} over random $10$-class subsets of the dataset vocabulary. ``Full Vocab.'' is the
dataset-level AP. ``Random'' is the mean$\pm$std over $5{,}000$ random $10$-class
subsets. ``Adverse'' and ``Favorable'' are the worst and best \emph{possible}
$10$-class subsets, taken as the bottom- and top-$10$ categories by per-category AP,
and they bound the achievable range. ``Fav./Full'' is the favorable-bound inflation
over the full-vocabulary number. \#CLS\,(GT) counts categories with at least one
ground-truth instance in the test set, fewer than the dataset-level $83$/$196$
classes.}
\label{tab:basenovel}
\vspace{-1em}
\end{table}

\noindent\textbf{Sensitivity to Class Split.} A widely used alternative protocol partitions categories into {\tt base} and {\tt novel} sets and reports AP on the {\tt novel} classes held out during training~\cite{lu2023ov3det,yang2024imov3d,huang2024ovm3ddet}. However, these numbers are not comparable across papers since they differ in which classes are withheld. We isolate the evaluation-side effect by re-scoring a frozen model that has seen all categories, reporting the mean $\pm$ std over $5{,}000$ random $10$-class subsets alongside the adverse and favorable bounds (bottom- and top-$10$ categories by per-category AP). As \cref{tab:basenovel} shows, reported AP depends strongly on which categories are scored, with DetAny3D at $8.7\pm3.9$ across random 10 class subsets on SUNRGBD. This further motivates our fixed dataset-level vocabulary.

\begin{wrapfigure}{r}{0.45\textwidth}
\centering
\includegraphics[width=0.85\linewidth]{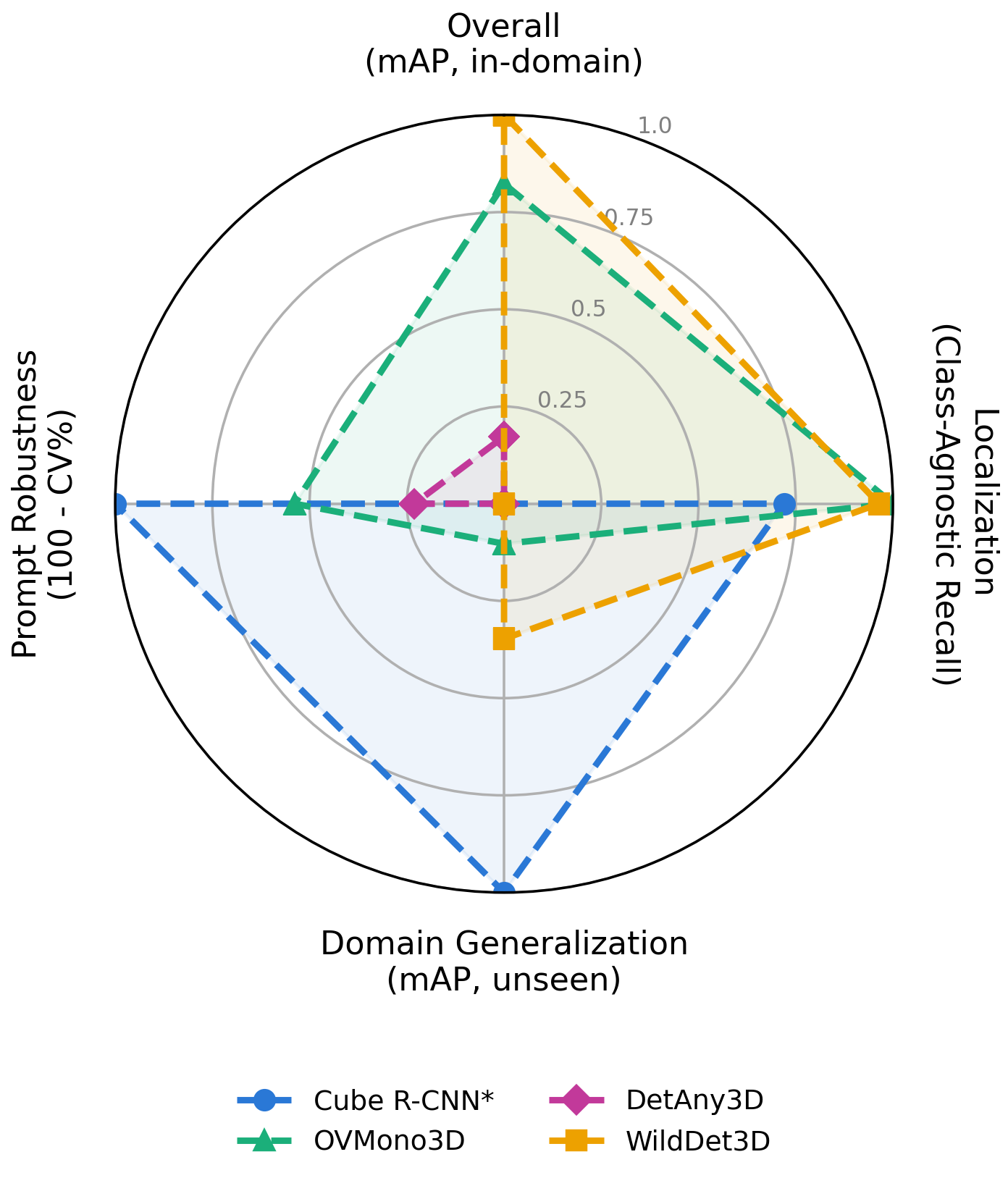}
\caption{\textbf{No Single Axis Predicts Another.} We compare monocular 3D detectors along three benchmark axes and mAP\textsubscript{3D}, min-max normalized across the four detectors. WildDet3D tops the in-domain axis yet ranks last on prompt robustness; Cube R-CNN* is weakest in-domain yet leads domain generalization and prompt robustness.}
\label{fig:radar}
\end{wrapfigure}

\newpage

\noindent\textbf{Breakdown Analysis.} The three axes we analyze in  \cref{fig:radar} expose trade-offs that a single mAP\textsubscript{3D} number conflates. Detector recall tracks dataset difficulty rather than clustering tightly (\cref{tab:tp_iou3d}), yet it consistently exceeds what mAP\textsubscript{3D} credits, so mAP gaps chiefly reflect semantic accuracy, not geometry. In-domain and out-of-domain accuracy disagree. DetAny3D and OVMono3D rank among the strongest in-domain, yet DetAny3D transfers worst out-of-domain, whereas our remapped Cube R-CNN$^*$ ranks weakest in-domain yet generalizes best outdoors. Prompt robustness is orthogonal to both axes. OVMono3D and WildDet3D lead the localization axis by a narrow margin, yet WildDet3D ranks among the least prompt-stable, and DetAny3D's top-line AP is vulnerable to descriptive prompts and target-aware inflation. No axis predicts another, so a method that advances on one can silently regress the rest.

\newpage 

\label{sec:runtime}
\begin{wraptable}{r}{0.44\textwidth}
\centering
\scriptsize
\setlength{\tabcolsep}{4pt}
\renewcommand{\arraystretch}{0.9}
\begin{tabular}{l|ccc}
\textbf{Method} & \textbf{Sem.} & \textbf{Geom.} & \textbf{Total} \\
\midrule
Cube R-CNN* & 190.6 & 33.3 & \textbf{223.9} \\
OVM3D-Det*  & 190.6 & 83.0 & 273.6 \\
OVMono3D    & \multicolumn{2}{c}{300.8} & 300.8 \\
DetAny3D    & 182.9 & 400.0 & 582.9
\end{tabular}
\caption{\textbf{End-to-End Latency (ms/image).} We benchmark runtime on 1 H100 by stage. Cube R-CNN* is fastest at $223.9$\,ms, $2.6\times$ ahead of DetAny3D.}
\label{tab:latency}
\end{wraptable}

\noindent\textbf{Runtime Analysis.} Our SigLIPv2 remapping encodes every detected crop, so we quantify this overhead against end-to-end open-vocabulary baselines on a single NVIDIA H100 NVL GPU over a fixed $200$-image ARKitScenes subset. Notably, every open-vocabulary baseline already relies on a heavyweight semantic model. OVMono3D~\cite{yao2025ovmono3d} and DetAny3D~\cite{zhang2025detany3d} both invoke GroundingDINO~\cite{groundingDINO}, which we include for a faithful comparison. \Cref{tab:latency} decomposes each pipeline into semantic and geometry stages. We find that our semantic stage costs $190.6$\,ms/image, matching the GroundingDINO stage that DetAny3D already requires ($182.9$\,ms/image). The difference lies in the geometric stage, where our closed-vocabulary detectors are cheap ($33.3$\,ms for Cube R-CNN, $83.0$\,ms for OVM3D-Det), whereas DetAny3D pays $400$\,ms for its SAM-based 3D head. Our full Cube R-CNN* pipeline is therefore the \emph{fastest} open-vocabulary detector here at $223.9$\,ms/image, $2.6\times$ ahead of DetAny3D, while remaining competitive in accuracy (\cref{tab:standard_metrics}). OVMono3D fuses both stages, running in $300.8$\,ms.

\noindent\textbf{Limitations.} The VLM remapping cannot cleanly separate true positives from false positives, since their cosine similarities overlap. We posit that lightweight calibration (\eg~an MLP on held-out similarity scores) would help. Further, OV3D-Bench inherits the annotation conventions and non-exhaustive labels of its constituent datasets, so a correct detection of an unannotated object is penalized as a false positive. Lastly, our remapping baseline is a diagnostic probe, not a deployable detector.

\section{Conclusion}
We introduce OV3D-Bench, a diagnostic benchmark for open-vocabulary monocular 3D detection that decouples geometric accuracy, semantic robustness, and cross-domain generalization under a dataset-level protocol. We benchmark seven detectors and introduce a simple training-free VLM remapping baseline that holds geometry fixed while varying only the vocabulary. We find that geometric localization is more mature, while semantic vocabulary remains the larger, unresolved gap that the community should prioritize closing. Under OV3D-Bench, a method reports three numbers: 3D Class-Agnostic Recall, dataset-level mAP\textsubscript{3D}, and prompt CV. For method developers, the gap between 3D Class-Agnostic Recall and mAP\textsubscript{3D} isolates whether a failure is geometric or semantic. For practitioners, the confusion matrix and prompt CV expose deployment behavior that current leaderboards hide.

\clearpage
\clearpage
{
    \bibliographystyle{splncs04}
    \bibliography{main}

\begin{thebibliography}{10}
\providecommand{\url}[1]{\texttt{#1}}
\providecommand{\urlprefix}{URL }
\providecommand{\doi}[1]{https://doi.org/#1}

\bibitem{brazil2023omni3d}
Brazil, G., Kumar, A., Straub, J., Ravi, N., Johnson, J., Gkioxari, G.:
  {Omni3D}: A large benchmark and model for {3D} object detection in the wild.
  In: CVPR. IEEE, Vancouver, Canada (June 2023)

\bibitem{caesar_nuscenes_2020}
Caesar, H., Bankiti, V., Lang, A.H., Vora, S., Liong, V.E., Xu, Q., Krishnan,
  A., Pan, Y., Baldan, G., Beijbom, O.: {nuScenes}: {A} multimodal dataset for
  autonomous driving. In: {CVPR} (2020)

\bibitem{carion2025sam3}
Carion, N., Gustafson, L., Hu, Y.T., Debnath, S., Hu, R., Suris, D., Ryali, C.,
  Alwala, K.V., Khedr, H., Huang, A., et~al.: Sam 3: Segment anything with
  concepts. arXiv preprint arXiv:2511.16719  (2025)

\bibitem{davidson2025refav}
Davidson, C., Ramanan, D., Peri, N.: Refav: Towards planning-centric scenario
  mining. arXiv preprint arXiv:2505.20981  (2025)

\bibitem{Dehghan2021arkitscenes}
Dehghan, A., Baruch, G., Chen, Z., Feigin, Y., Fu, P., Gebauer, T., Kurz, D.,
  Dimry, T., Joffe, B., Schwartz, A., Shulman, E.: Arkitscenes: A diverse
  real-world dataset for 3d indoor scene understanding using mobile rgb-d data.
  In: NeurIPS Datasets and Benchmarks (2021)

\bibitem{geiger2012kitti}
Geiger, A., Lenz, P., Urtasun, R.: Are we ready for autonomous driving? the
  kitti vision benchmark suite. In: 2012 IEEE conference on computer vision and
  pattern recognition. pp. 3354--3361. IEEE (2012)

\bibitem{huang2024ovm3ddet}
Huang, R., Zheng, H., Wang, Y., Xia, Z., Pavone, M., Huang, G.: Training an
  open-vocabulary monocular 3d detection model without 3d data. In: Advances in
  Neural Information Processing Systems. vol.~37, pp. 72145--72169. Curran
  Associates, Inc. (2024)

\bibitem{huang2026wilddet3d}
Huang, W., Zhang, J., Li, S., Jia, T., et~al.: Wilddet3d: Scaling promptable 3d
  detection in the wild. arXiv preprint arXiv:2604.08626  (2026)

\bibitem{khurana2024shelf}
Khurana, M., Peri, N., Hays, J., Ramanan, D.: Shelf-supervised cross-modal
  pre-training for 3d object detection. arXiv preprint arXiv:2406.10115  (2024)

\bibitem{li2025modelingvariantsprompts}
Li, A., Liu, Z., Li, X., Zhang, J., Wang, P., Wang, H.: Modeling variants of
  prompts for vision-language models (2025)

\bibitem{lin2026userpromptingstrategiesprompt}
Lin, J., Xiu, Y., Gorlatova, M.: User prompting strategies and prompt
  enhancement methods for open-set object detection in xr environments (2026)

\bibitem{groundingDINO}
Liu, S., Zeng, Z., Ren, T., Li, F., Zhang, H., Yang, J., Jiang, Q., Li, C.,
  Yang, J., Su, H., Zhu, J., Zhang, L.: Grounding dino: Marrying dino with
  grounded pre-training for open-set object detection. In: Computer Vision --
  ECCV 2024: 18th European Conference, Milan, Italy, September 29--October 4,
  2024, Proceedings, Part XLVII. pp. 38--55. Springer-Verlag, Berlin,
  Heidelberg (2024)

\bibitem{lu2023ov3det}
Lu, Y., Xu, C., Wei, X., Xie, X., Tomizuka, M., Keutzer, K., Zhang, S.:
  Open-vocabulary point-cloud object detection without 3d annotation. In: 2023
  IEEE/CVF Conference on Computer Vision and Pattern Recognition (CVPR). pp.
  1190--1199. IEEE (2023)

\bibitem{ma2025longtailed3ddetectionmultimodal}
Ma, Y., Peri, N., Dave, A., Hua, W., Ramanan, D., Kong, S.: Long-tailed 3d
  detection via multi-modal fusion (2025)

\bibitem{madan_revisiting_2024}
Madan, A., Peri, N., Kong, S., Ramanan, D.: Revisiting few-shot object
  detection with vision-language models. In: Advances in Neural Information
  Processing Systems. vol.~37, pp. 19547--19560. Curran Associates, Inc. (2024)

\bibitem{ovsep2024better}
O{\v{s}}ep, A., Meinhardt, T., Ferroni, F., Peri, N., Ramanan, D.,
  Leal-Taix{\'e}, L.: Better call sal: Towards learning to segment anything in
  lidar. In: European Conference on Computer Vision. pp. 71--90. Springer
  (2024)

\bibitem{peri2022towards}
Peri, N., Dave, A., Ramanan, D., Kong, S.: Towards long tailed 3d detection.
  CoRL  (2022)

\bibitem{intelligence2025pi}
{Physical Intelligence}, Black, K., Brown, N., Darpinian, J., Dhabalia, K.,
  Driess, D., Esmail, A., Equi, M., Finn, C., Fusai, N., et~al.: Pi 0.5: a
  vision-language-action model with open-world generalization. arXiv preprint
  arXiv:2504.16054  (2025)

\bibitem{piccinelli2024unidepth}
Piccinelli, L., Yang, Y.H., Sakaridis, C., Segu, M., Li, S., Van~Gool, L., Yu,
  F.: Unidepth: Universal monocular metric depth estimation. In: Proceedings of
  the IEEE/CVF Conference on Computer Vision and Pattern Recognition. pp.
  10106--10116 (2024)

\bibitem{radford2021clip}
Radford, A., Kim, J.W., Hallacy, C., Ramesh, A., Goh, G., Agarwal, S., Sastry,
  G., Askell, A., Mishkin, P., Clark, J., et~al.: Learning transferable visual
  models from natural language supervision. In: International conference on
  machine learning. pp. 8748--8763. PmLR (2021)

\bibitem{ren2024gsam}
Ren, T., Liu, S., Zeng, A., Lin, J., Li, K., Cao, H., Chen, J., Huang, X.,
  Chen, Y., Yan, F., et~al.: Grounded sam: Assembling open-world models for
  diverse visual tasks. arXiv preprint arXiv:2401.14159  (2024)

\bibitem{Roberts2020Hypersim}
Roberts, M., Paczan, N.: Hypersim: A photorealistic synthetic dataset for
  holistic indoor scene understanding. 2021 IEEE/CVF International Conference
  on Computer Vision (ICCV) pp. 10892--10902 (2020)

\bibitem{robicheaux2025roboflow100vl}
Robicheaux, P., Popov, M., Madan, A., Robinson, I., Nelson, J., Ramanan, D.,
  Peri, N.: Roboflow100-vl: A multi-domain object detection benchmark for
  vision-language models. Advances in Neural Information Processing Systems
  (2025)

\bibitem{rozenberszki2022language}
Rozenberszki, D., Litany, O., Dai, A.: Language-grounded indoor 3d semantic
  segmentation in the wild. In: European Conference on Computer Vision (ECCV)
  (2022)

\bibitem{song2015sunrgbd}
Song, S., Lichtenberg, S.P., Xiao, J.: Sun rgb-d: A rgb-d scene understanding
  benchmark suite. In: Proceedings of the IEEE conference on computer vision
  and pattern recognition. pp. 567--576 (2015)

\bibitem{takmaz2025towards}
Takmaz, A., Saltori, C., Peri, N., Meinhardt, T., De~Lutio, R., Leal-Taix{\'e},
  L., O{\v{s}}ep, A.: Towards learning to complete anything in lidar. arXiv
  preprint arXiv:2504.12264  (2025)

\bibitem{tschannen2025siglip2}
Tschannen, M., Gritsenko, A., Wang, X., Naeem, M.F., Alabdulmohsin, I.,
  Parthasarathy, N., Evans, T., Beyer, L., Xia, Y., Mustafa, B., et~al.: Siglip
  2: Multilingual vision-language encoders with improved semantic
  understanding, localization, and dense features. arXiv preprint
  arXiv:2502.14786  (2025)

\bibitem{wen2024foundationpose}
Wen, B., Yang, W., Kautz, J., Birchfield, S.: Foundationpose: Unified 6d pose
  estimation and tracking of novel objects. In: Proceedings of the IEEE/CVF
  conference on computer vision and pattern recognition. pp. 17868--17879
  (2024)

\bibitem{Wilson2021Argoverse2}
Wilson, B., Qi, W., Agarwal, T., Lambert, J., Singh, J., Khandelwal, S., Pan,
  B., Kumar, R., Hartnett, A., Pontes, J.K., Ramanan, D., Carr, P., Hays, J.:
  Argoverse 2: Next generation datasets for self-driving perception and
  forecasting. In: Proceedings of the Neural Information Processing Systems
  Track on Datasets and Benchmarks (NeurIPS Datasets and Benchmarks 2021)
  (2021)

\bibitem{yang2024imov3d}
Yang, T., Ju, Y., Yi, L.: Imov3d: Learning open vocabulary point clouds 3d
  object detection from only 2d images. Advances in Neural Information
  Processing Systems  \textbf{37},  141261--141291 (2024)

\bibitem{yao2025ovmono3d}
Yao, J., Gu, H., Chen, X., Wang, J., Cheng, Z.: Open vocabulary monocular {3D}
  object detection (2024), arXiv:2411.16833

\bibitem{zhang2025detany3d}
Zhang, H., Jiang, H., Yao, Q., Sun, Y., Zhang, R., Zhao, H., Li, H., Zhu, H.,
  Yang, Z.: Detect anything 3d in the wild. In: Proceedings of the IEEE/CVF
  International Conference on Computer Vision (ICCV). pp. 5048--5059 (October
  2025)

\end{thebibliography}
}

\end{document}